\documentclass[10pt,twocolumn,letterpaper]{article}

\usepackage[pagenumbers]{cvpr} %

\usepackage{algorithm}
\usepackage{algpseudocode}
\usepackage{multirow}
\usepackage{booktabs}
\usepackage{amssymb}
\usepackage{amsmath}
\usepackage{colortbl, array, xcolor}

\definecolor{cvprblue}{rgb}{0.21,0.49,0.74}
\usepackage[pagebackref,breaklinks,colorlinks,allcolors=cvprblue]{hyperref}

\usepackage{lipsum}
\usepackage[accsupp]{axessibility} %
\newcommand{\stdk}[1]{\footnotesize $\pm$#1}
\newcommand{\std}[1]{}
\newcommand{\oursfull}[1]{Type-Retouch}
\newcommand{\ours}[1]{Type-R}

\RequirePackage[breakable,skins]{tcolorbox}
\usepackage{colortbl}
\definecolor{colorcommentfg}{RGB}{0,63,87}
\definecolor{colorcommentbg}{HTML}{CCCCFF}
\definecolor{colorcommentframe}{HTML}{76608A}

\newenvironment{intentionpromptaug}[1][]{
	\begin{tcolorbox}[adjusted title={System message for augmentation}, fonttitle={\bfseries\footnotesize}, fontupper=\scriptsize, colback={colorcommentbg!40}, colframe={colorcommentframe!90},coltitle={white},#1]
}{\end{tcolorbox}}

\newenvironment{intentionpromptregeneration}[1][]{
	\begin{tcolorbox}[adjusted title={System message for regeneration}, fonttitle={\bfseries\footnotesize}, fontupper=\scriptsize, colback={colorcommentbg!40}, colframe={colorcommentframe!90},coltitle={white},#1]
}{\end{tcolorbox}}

\newenvironment{intentionprompt}[1][]{
	\begin{tcolorbox}[adjusted title={System message for evaluation}, fonttitle={\bfseries\footnotesize}, fontupper=\scriptsize, colback={colorcommentbg!40}, colframe={colorcommentframe!90},coltitle={white},#1]
}{\end{tcolorbox}}

\def\paperID{2486} %
\def\confName{CVPR}
\def\confYear{2025}

\title{Type-R: Automatically Retouching Typos for Text-to-Image Generation}

\author{
    Wataru Shimoda\textsuperscript{1} \quad Naoto Inoue\textsuperscript{1} \quad Daichi Haraguchi\textsuperscript{1} \\ 
    Hayato Mitani\textsuperscript{2} \quad Seiichi Uchida\textsuperscript{2} \quad Kota Yamaguchi\textsuperscript{1} \\
    $^{1}$CyberAgent \quad ${^2}$Kyushu University, Japan
}

\begin{document}

\twocolumn[{%
\renewcommand\twocolumn[1][]{#1}%
\maketitle
\begin{center}
    \centering
    \captionsetup{type=figure}
    \includegraphics[width=\textwidth]{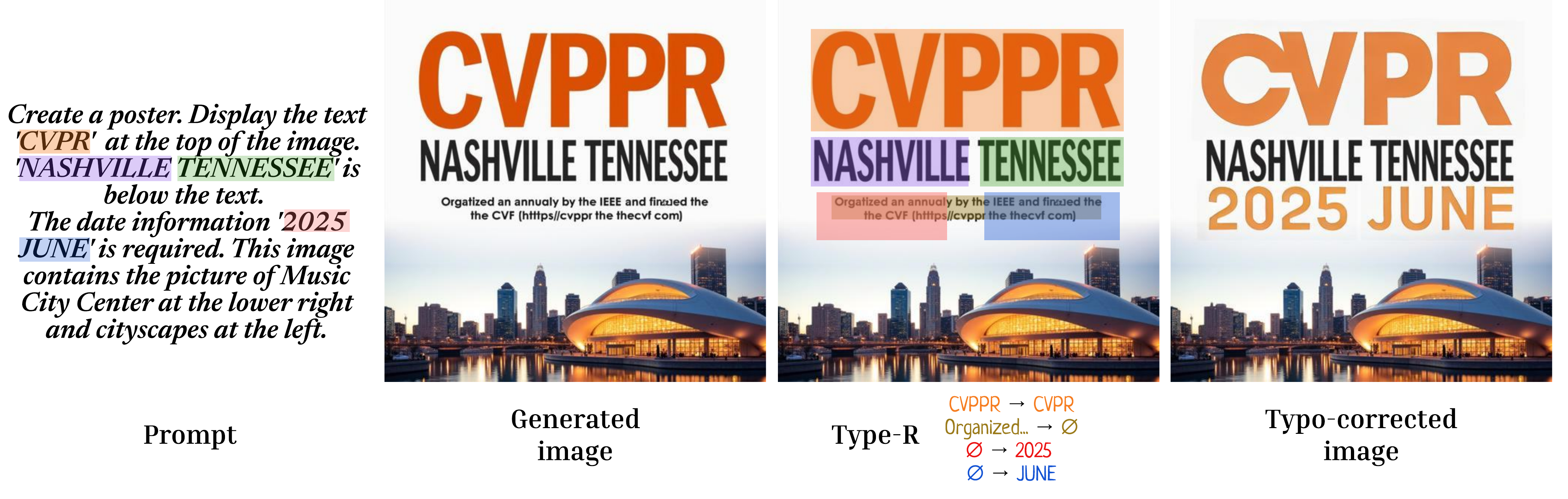}
    \caption{\ours{} automatically corrects typographic errors in an image generated from any text-to-image model without sacrificing their visual quality. We highlight prompt words corresponding to the words in the image with the same color.}
    \label{fig:teaser}
\end{center}%
}]

\maketitle
\begin{abstract}
While recent text-to-image models can generate photorealistic images from text prompts that reflect detailed instructions, they still face significant challenges in accurately rendering words in the image.
In this paper, we propose to retouch erroneous text renderings in the post-processing pipeline.
Our approach, called \ours{}, identifies typographical errors in the generated image, erases the erroneous text, regenerates text boxes for missing words, and finally corrects typos in the rendered words.
Through extensive experiments, we show that \ours{}, in combination with the latest text-to-image models such as Stable Diffusion or Flux, achieves the highest text rendering accuracy while maintaining image quality and also outperforms text-focused generation baselines in terms of balancing text accuracy and image quality.
\footnote{\href{https://github.com/CyberAgentAILab/Type-R}{Code} and \href{https://cyberagentailab.github.io/Type-R}{project page} are available.}

\end{abstract}

\section{Introduction} \label{sec:introduction}
The rapid advancement of text-to-image generation models triggers a surge of application interest in various creative domains.
Graphic design is one of the domains from which text-to-image models potentially benefit, with the goal of helping generate appealing posters, advertisements, book covers, webpages, or signs, such that these display media convey the intended message to the viewers.
However, one of the major obstacles is text rendering on the image~\cite{liu2022character}.
Popular state-of-the-art text-to-image models such as Stable Diffusion~\cite{podell2023sdxl,sd3dot5} or Flux~\cite{flux1dev} still face notable challenges in accurately rendering words or characters (\cref{fig:teaser} second column).
Common issues include missing words, unintended characters, or word-level errors that insert another similar-looking word.
While it might be possible to tune the text-to-image models with increasing amounts of training data, the resulting image is not guaranteed to be typo-free.

In this work, instead of modifying the image generator, we propose automatically retouching the erroneous text renderings of the already generated image.
Our approach, named \ours{}, detects typographic errors in a generated image and automatically fixes erroneous content.
\Cref{fig:teaser} shows an example of the input prompt, the generated image by {Flux.1-dev}~\cite{flux1dev}, detected errors by \ours{}, and the final retouched image.
\ours{} successfully corrects the typographic errors without artifacts compared to the initial generation.
Our approach is fully automatic, independent of the base text-to-image generator, and easily pluggable to any existing text-to-image model.

Through extensive experiments, we study the limitations in the current text-to-image models, including both the popular text-to-image generators~\cite{podell2023sdxl,flux1dev} and typography-focused models~\cite{textdiffuser,textdiffuser2}, and show that \ours{} significantly improves the text rendering accuracy of text-to-image models while maintaining graphic design quality.
We also study several design choices, \eg, how changing the base text-to-image generator or the OCR model affects the resulting visual quality and text accuracy, and show that \ours{} in combination with Flux~\cite{flux1dev} achieves the best trade-off between text rendering accuracy and quality scores by GPT evaluation on Mario-Eval benchmark~\cite{textdiffuser}.

We summarize our contributions in the following
\begin{enumerate}
  \item We propose \ours{}, which automatically retouches typographic errors in generated images by text-to-image models while maintaining the design quality.
  \item We empirically show that the combination of the state-of-the-art text-to-image generator and \ours{} achieves the best quality-accuracy trade-off and outperforms the approach specifically designed for typography.
\end{enumerate}

\begin{figure*}[t]
\centering
\includegraphics[width=\hsize]{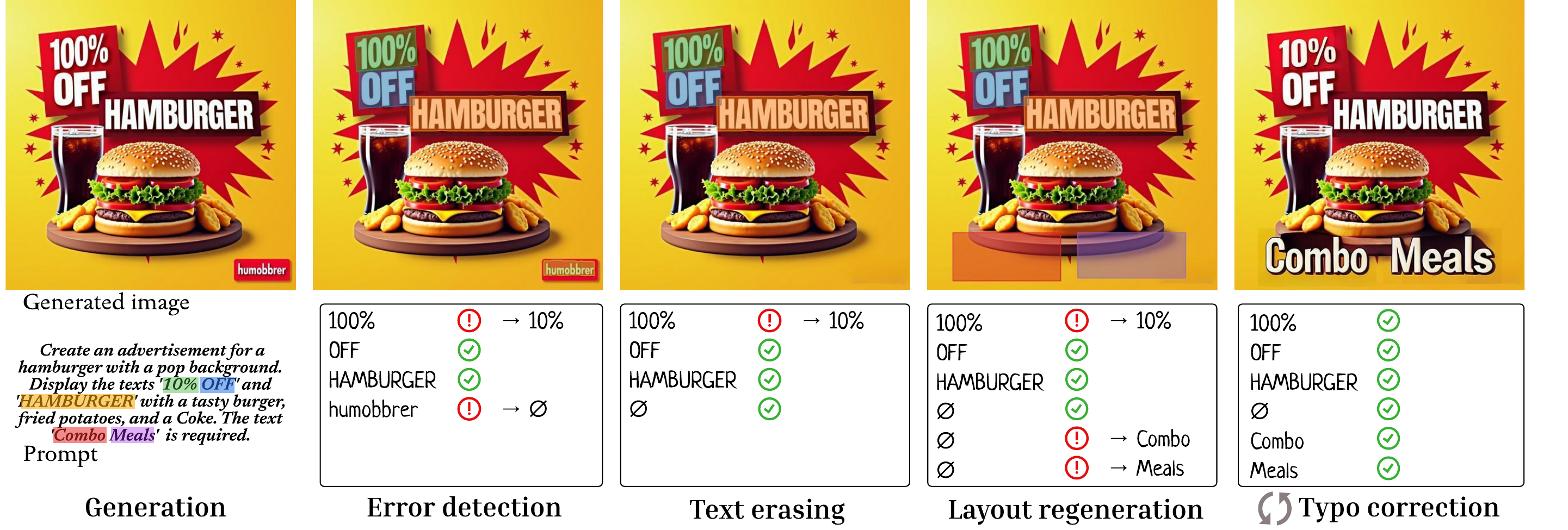}
\caption{Illustration of the \ours{} pipeline. \ours{} automatically detects errors, erases unintended texts, inserts missing words, and corrects spelling errors in the image.}
\label{fig:pipeline}
\end{figure*}

\section{Related Work} \label{sec:related_work}

\subsection{Text-to-Image Generation} \label{sec:text-to-image}
Text-to-image generation is the task of creating images from text prompts.
With the development of large image-caption datasets, there has been significant progress in generating photorealistic images from diffusion-based models~\cite{rombach2022high, podell2023sdxl, dalle3,saharia2022photorealistic, deepfloyd_if}.
However, rendering accurate text is still one of the remaining problems.
Liu \etal~\cite{liu2022character} point out the challenge in accurately rendering characters in text-to-image generation models, where it is common to observe missing glyphs, redundant glyphs, or wrong characters.
The text accuracy problem persists in the recent text-to-image models such as Stable Diffusion 3~\cite{esser2024scaling} or Flux~\cite{flux1dev}.

In this work, instead of modifying or tuning the text-to-image models, we utilize feedback from external models to improve text rendering.
Several works report the successful use of external models in improving image generation.
RPG~\cite{yang2024mastering} introduces a technique for region-specific prompt guidance for noise control. 
Wu \etal~\cite{wu2024self} apply object detection to generated images and refine detection results using Large Language Models.
In our task setup, we take advantage of pre-trained Vision Language Models (VLMs), OCR models, and text-inpainting models to retouch typographic errors.

\subsection{Visual Text Generation} \label{sec:visual_text_generation}

Several text-to-image models~\cite{deepfloyd_if, saharia2022photorealistic, zhangli2024layout, liu2024glyph, liu2024glyphv2} have reported that better text encoder significantly improves text rendering capabilities, where the encoder can learn glyph-level information from the training corpus.
Adding OCR losses also improve performances~\cite{li2024empowering, anytext, zhao2025udifftext}.
While improving the capability of the base text-to-image generator might eventually reduce typographic errors, we consider a post-processing approach independent of the base model.

Another line of work to improve text rendering is via conditional inputs to the text-to-image models.
Some literature attempts at conditioning generation on text positions and glyph masks to improve the text rendering accuracies~\cite{yang2024glyphcontrol, ma2023glyphdraw, anytext, zhang2024brush, typographyctl, textdiffuser, textdiffuser2, textharmony}.
Particularly, TextDiffusers~\cite{textdiffuser,textdiffuser2} generate images with two steps: layout planning, which generates text position information from prompts, followed by image generation conditioned on the first layout information.
TextDiffuser~\cite{textdiffuser} and TextDiffuser 2~\cite{textdiffuser} differ in how they generate layout information, where the former generates character-by-character masks and the latter generates layout in language.
One major limitation of conditional generation is the training requirements; \ie, building a sufficient training corpus for the conditioning module is often difficult.
Our experiments suggest that TextDiffusers suffer from inadequate image quality compared to the state-of-the-art image generators, likely due to the bias in its training data.

We take the approach of retouching afterward using external pre-trained models to take full advantage of the best text-to-image models without fine-tuning.
A few works incorporate external models for text rendering.
Very recently, Lakhanpal \etal reports Refining T2I~\cite{lakhanpal2024refining} that applies OCR to the generated images by TextDiffuser and optimizes the layouts to improve the legibility of texts, while the same limitation to TextDiffusers still applies.
Li \etal~\cite{Li2024first} propose to generate background images by removing generated scene text.
Unlike prior works, we propose a method for generating images that adheres as much as possible to the initial generation results, making full use of OCR models, text removal models, and VLMs to improve text rendering.

\subsection{Text Editing and Synthesis}
Another relevant task is raster text editing, which directly manipulates texts in the image.
While GAN-based methods~\cite{srnet,roy2020stefann, yang2020swaptext,mostel} have been popular, recent visual text generation methods~\cite{diffste, anytext, wang2024high, wang2024textmaster, li2024joytype, vtgw, zeng2024textctrl} are capable of editing texts in the image directly.
While text editing and our task setup seem similar, text editing is only a part of the typography retouching task, which involves further tasks such as automatic error detection or missing word insertion.

\section{Approach} \label{sec:approach}
Our \ours{} pipeline consists of four stages, which are all automatic: 1) error detection, 2) text erasing, 3) layout regeneration, and 4) typo correction.
We show the pipeline overview in \cref{fig:pipeline}.

\subsection{Error Detection}

We denote an image $I$ generated by an off-the-shelf text-to-image model $G$ with a prompt, 
which requests to show $N$ words $W = \{w_{i}\}_{i=1}^N$ in $I$.
To ensure how much this expectation is satisfied, we first detect words in the image $I$.
Specifically, a scene text detection model predicts polygonal word regions $\hat{B}$.
We remove polygons with a height less than a threshold $\theta$.
Then, a scene text recognition model parses the words $\hat{W}= \{\hat{w}_{i}\}_{i=1}^{\hat{N}}$ in the regions $\hat{B}$.
As noted in Section~\ref{sec:introduction}, the model $G$ is often not perfect, and $W\neq \hat{W}$; therefore, we need to identify errors in $\hat{W}$ by considering $W$ is a ground-truth.
\par

Roughly speaking, this identification task is a matching problem between $W$ and 
$\hat{W}$. More specifically, we consider the matching between the word set $\mathcal{W}$ and $\hat{\mathcal{W}}$ of equal size $\mathcal{N}$ by optimal transport:
\begin{equation}
   \textup{OT}(\mathcal{W}, \hat{\mathcal{W}}) \equiv \underset{\pi}{\mathrm{max}} \sum_{i = 1}^{\mathcal{N}}\textup{Lev}(w_{i}, \hat{w}_{\pi(i)}), 
\end{equation}
where $w_{i} \in \mathcal{W}, \hat{w}_{i} \in \hat{\mathcal{W}}$, $\pi$ is a one-by-one matching, and $\textup{Lev}(\cdot,\cdot)$ is the Levenshtein distance to measure the text similarity.
This matching works only when the two sets are of equal size, but typographic errors result in different sizes because of missing or unintended words.
Here, we equalize two word sets $W$ and $\hat{W}$ by padding tokens.
Let us define $\mathcal{N} = \mathrm{max} (N, \hat{N})$.
We express the padded word sets by:
\begin{align}
    \mathcal{W} = W \cup \{ p_i \}_{i=N+1}^\mathcal{N}, \\
    \hat{\mathcal{W}} = \hat{W} \cup \{ \hat{p}_i \}_{i=\hat{N}+1}^\mathcal{N},
\end{align}
where $p_i$ and $\hat{p}_i$ are the padding tokens.
Here, a word matching a padding token indicates a typographic error; \ie, a paired word with $p$ is an unintended word, and a paired word with $\hat{p}$ is a missing word.
We assign a constant distance to any word matching to the padding tokens.
Any pairs having non-zero distance indicate spelling errors.

The following sections describe our approach to minimizing the above optimal transportation cost.
We align the number of words by erasing redundant texts when $N < \hat{N}$ (\cref{sec:texterasing}) and insert text boxes by layout regeneration when $N > \hat{N}$ (\cref{sec:layoutregneration}).
Finally, we fix spelling errors in typo correction (\cref{sec:typo_correction}).

\subsection{Text Erasing}
\label{sec:texterasing}
\ours{} erases unintended texts using an off-the-shelf inpainting method~\cite{suvorov2022resolution} if the number of detected words exceeds the number of the words specified by a prompt, i.e., $N<\hat{N}$, or the generated image includes filtered texts.
This step reduces the number of words in a generated image to equal or less than the number of specified words by the prompt.
The inpainting model removes specified paint regions.
In our setup, we specify regions that we want to erase in $\hat{B}$ and composite the inpainted results to the original image.
Since text detection may not always fully cover text areas, we slightly enlarge the text region masks to ensure complete removal.

\subsection{Layout Regeneration}
\label{sec:layoutregneration}

Layout regeneration fills in missing text regions so that the resulting $\hat{N}$ equals the number of words specified by a prompt $N$ when $\hat{N} < N$.
This is equivalent to obtaining position information from the image for missing words of the prompt, and we address this task using a VLM, specifically GPT-4o~\cite{gpt4o}.
Inspired by works on layout generation by zero/few-shot learning of LLMs~\cite{lin2023layoutprompter,feng2023layoutgpt}, we ask GPT to plan layouts for each missing word from the image and layout information of valid text boxes in JSON format.

For simplicity, we represent the layout information as a bounding box with four coordinates. %
When text detection outputs are in the polygonal form, we convert them to boxes by finding the smallest four-sided bounding box that encloses each polygon.

\subsection{Typo Correction}
\label{sec:typo_correction}
Typo correction tries to resolve all word pairs with non-zero edit distance.
To address word-level spelling errors, we employ a pre-trained text editing model~\cite{anytext} to automatically retouch the word region.
Using an inpainting approach, the text editing model generates raster images containing specified text in the selected areas.
Similar to text-to-image models, text editing models may not always render the specified text accurately but with a much lower computation cost than large text-to-image models.
We repeatedly apply the text editing model until we fix the spelling errors or we reach the predefined number of attempts.
Note that we only composite the areas of corrected words in the edited image during the iteration.
We show the pseudo-codes in \cref{alg:tmp}.

\begin{algorithm}
\caption{A pseudo-code of typo correction. }
\label{alg:tmp}
\begin{algorithmic}[1]
\Function {Typo correction}{Image $ I$, Words $W$, Detected regions $\hat{B}$, Maximum iterations $t_\mathrm{max}$}
    \For{$t=1$ to $t_{\mathrm{max}}$}
        \State{$I_{\mathrm{edit}}$ $\leftarrow$ \textup{Text-Editor}$(I, W, \hat{B})$} 
        \State{$\hat{W}$ $\leftarrow$ \textup{Text-Recognition}$(I_{\mathrm{edit}}, \hat{B})$} 
        \State{ $\mathcal{O}$  $\leftarrow$ \textup{Diff}$(W, \hat{W})$} \Comment{ Find spelling errors}
        \State{$I$ $\leftarrow$ \textup{Composite}$(I, I_{\mathrm{edit}}, \hat{B}, \mathcal{O})$} 
        \If{$\mathcal{O} \neq \varnothing$}
            \State{$W, \hat{B}$  $\leftarrow$ \textup{Update}$(W, \hat{B}, \mathcal{O})$} 
        \Else
            \State{ break}         
        \EndIf
    \EndFor
    \State{\textbf{return} $I$}
\EndFunction
\end{algorithmic}
\end{algorithm}

\section{Experiment}

\begin{figure*}[t]
\centering
\includegraphics[width=.9\hsize]{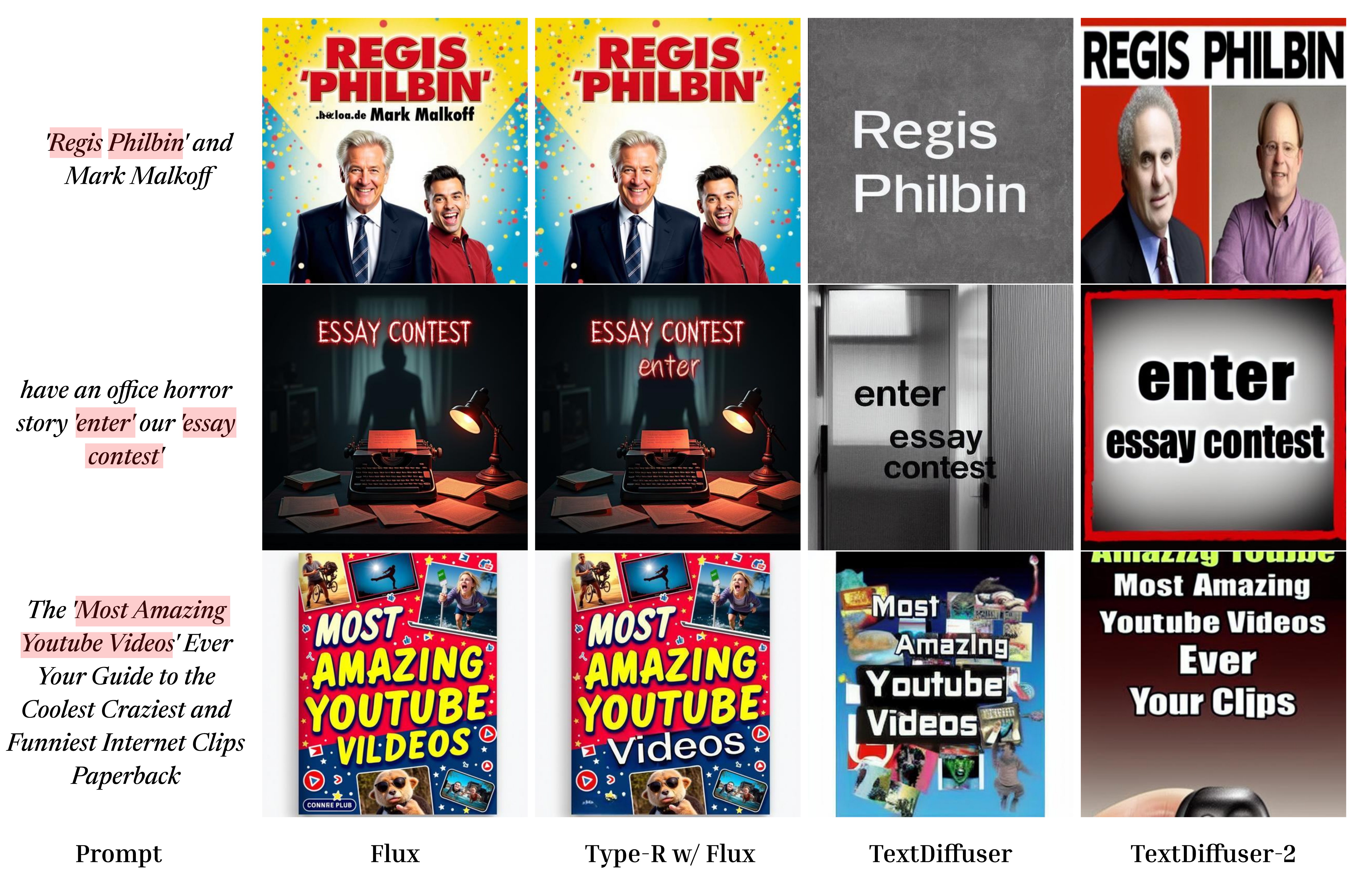}
\caption{Comparisons of generated images by Flux, Flux w/ \ours{}, and TextDiffuser models. The left column shows prompts, and the right columns present generated images by each method.}
\label{fig:cmp}
\end{figure*}

\subsection{Dataset}
\label{sec:dataset}
We evaluate the performance of visual text generation on MARIO-Eval benchmark~\cite{textdiffuser}.
This benchmark contains 5,414 prompts, of which 5,000 have image data.
We use all 5,414 prompts in the main comparison, i.e., in \cref{tab:main}. 
We randomly pick 500 of the 5,414 prompts as the subset for analysis in \cref{sec:analysis} because our experiments involve paid APIs.
For the selection of hyperparameters in filtering, we use 500 prompts randomly selected from the MARIO-LAION dataset~\cite{textdiffuser} as the validation set.  Note that the validation does not overlap with the benchmark.

\subsection{Metrics}
We evaluate the quality of generated images using a GPT-based evaluation similar to TextDiffuser-2~\cite{textdiffuser2}. 
In our evaluation, we adopt a rating-based evaluation approach~\cite{jia2023cole} instead of the voting evaluation because GPT-based voting is sensitive to the option order and the number of options.
In our settings, GPT assigns scores from 1 to 10 for both graphic design quality and content alignment between generated images and prompts.
For text rendering accuracy, we measure OCR accuracy commonly used for assessing text rendering performance used in text-to-image generation~\cite{textdiffuser,textdiffuser2}.
We also measure Fréchet Inception Distance (FID)~\cite{fid} to quantify distribution differences between generated images and the full 5000 images in the MARIO-Eval benchmark, and CLIP score~\cite{clip} to compute the similarity between each image and its prompt-based on the CLIP.

\subsection{Implementation Details}
We use multiple external models in our \ours{} pipeline without fine-tuning.
We use Deepsolo~\cite{deepsolo} for text region detection, Paddle~\cite{paddle} for scene text recognition, LaMa~\cite{touvron2023llama} for text erasing, GPT-4o~\cite{gpt4o} for layout regeneration, and AnyText~\cite{anytext} for text editing.
In addition, we use GPT to augment prompts following~\cite{white2023prompt}.
When augmenting prompts with GPT, we specify what categories of images to generate and how to draw the text in quotation marks to make it stand out.
For all GPT utilization in this paper, i.e., augmentation, layout planning, and evaluation, we use the same version of OpenAI API 2023-05-15.

When the response of the API is corrupted, we retry requests up to five times and give up.
For typo correction iteration, we set the maximum number of correction attempts $t_{\mathrm{max}}$ to 10.
We chose the hyperparameter $\theta$ for filtering to 4\% of the overall image height through grid search.

\subsection{Baselines} \label{sec:baselines}
Since \ours{} is a post-processing module, we combine \ours{} with the two base text-to-image models to build a baseline. In addition, we chose a few different baselines tailored for text rendering.

\noindent\textbf{\ours{} w/ SD3} combines Stable Diffusion 3~\cite{esser2024scaling} and our approach.

\noindent\textbf{\ours{} w/ Flux} combines Flux~\cite{flux1dev} and our approach.

\noindent\textbf{TextDiffuser~\cite{textdiffuser}} is a text-to-image generation method based on layout planning, which generates character-wise masks and generates images using the masks as conditions.

\noindent\textbf{TextDiffuser-2~\cite{textdiffuser2}} is a state-of-the-art text-to-image generation method based on layout planning. 
Unlike TextDiffuser, TextDiffuser-2 plans layouts via LLM and generates images by embedding the layout information into prompts.

\subsection{Results}
\paragraph{\ours{} vs. TextDiffusers}

\Cref{tab:main} summarizes evaluation metrics of TextDiffuser models and \ours{}.
The result suggests that \ours{} significantly outperforms the TextDiffuser models in GPT evaluation scores, and also \ours{} achieves higher content alignment.
The higher GPT scores indicate better aesthetic quality, as also suggested in the qualitative results in \Cref{fig:cmp}.
While the TextDiffuser models tend to generate images with simple backgrounds and texts (i.e., low quality in terms of graphic designs), \ours{} successfully generates images with contextually rich backgrounds and seamlessly integrated text through precise and corrective adjustments.

In OCR evaluations, SD3 with \ours{} yields lower scores than the TextDiffuser models, whereas Flux with \ours{} achieves higher accuracy. This result suggests that the enhancement of \ours{} is influenced by the capability of the base text-to-image model.

\begin{table}[t]
  \centering
\caption{Evaluation metrics for \ours{} and TextDiffuser baselines. ``Graphic'' and ``Match'' indicate graphic design quality and content alignment.} \label{tab:main}
  \begin{tabular}{@{}lccccc @{}}
    \hline
    \multirow{2}{*}{Method}
    & \multicolumn{2}{c}{GPT $\uparrow$}  
    & \multirow{2}{*}{OCR $\uparrow$} 
    & \multirow{2}{*}{FID$\downarrow$}  
    & \multirow{2}{*}{CLIP $\uparrow$} 
    \\
    & Graphic
    & Match
    & 
    & 
    \\
    \hline

 TD~\cite{textdiffuser} & 4.67\std{1.51} & 7.16\std{2.22} & 55.4 & 42.0 & 34.4  \\ %
 TD-2~\cite{textdiffuser2}  & 4.97\std{1.40}& 7.17\std{2.26}& 56.2 &\textbf{33.8} & \textbf{34.7} \\ %
 \rowcolor{gray!10}
 Type-R& &&  & & \\
 \rowcolor{gray!10}
 ~ SD3\cite{esser2024scaling} & 7.30\std{1.28}& 8.46\std{1.48} & 48.6 & 45.0 &33.0\\ %
 \rowcolor{gray!10}
 ~ Flux\cite{flux1dev} & \textbf{7.67}\std{1.25} & \textbf{8.55}\std{1.42} & \textbf{62.0} & 43.1 & 33.1 \\
    \hline
  \end{tabular}
\end{table}

\paragraph{User Study}

We also conducted a user study in which participants rated images using the same questions and scoring criteria as the GPT evaluations to validate the reliability of the GPT evaluation. 
We prepared 100 prompts and generated 100 corresponding images, which resulted in scores from 5 participants for a total of 500 ratings per method.

\Cref{tab:userstudy} presents the user study results. 
Both SD3 and Flux with \ours{} have higher scores than the TextDiffuser models, consistent with the GPT evaluation.
The differences in scores are small among all baselines, although we observe that all scores of \ours{} are statistically significant compared to TextDiffuser baselines.
The p-value of the paired t-test with \ours{} w/ Flux is $1.55e-11$ for TextDiffuser and $4.14e-18$ for TextDiffuser 2 in text-image matching evaluation.
The small difference might be due to a lack of confidence in rating design or the ability to perceive finer defects in graphic design.

\begin{table}[t]
  \centering
\caption{Userstudy for reliability of GPT evaluation.} \label{tab:userstudy}
  \begin{tabular}{@{}lcc@{}}
    \hline
    \multirow{2}{*}{Method} & \multicolumn{2}{c}{Human $\uparrow$} \\
    & Graphic &  Match\\ 
    \hline
    TextDiffuser~\cite{textdiffuser}     & 5.24 \stdk{2.18}          & 5.08 \stdk{2.51}\\
    TextDiffuser-2~\cite{textdiffuser2}  & 5.30 \stdk{1.60}          & 5.06 \stdk{2.39}\\
     \rowcolor{gray!10}
    Type-R & & \\
    \rowcolor{gray!10}
    ~ SD3\cite{esser2024scaling}           & 5.94 \stdk{1.86}          & 5.99 \stdk{2.34}\\
    \rowcolor{gray!10}
    ~ Flux\cite{flux1dev}                  & \textbf{6.23} \stdk{1.81} & \textbf{6.04} \stdk{2.34}\\
    \hline
  \end{tabular}
\end{table}

\paragraph{FID and CLIP scores}

Although TextDiffuser yields higher FID and CLIP scores, we argue that these scores may not accurately capture human perception of image quality.
For FID, the reference dataset is limited to 5,000 images that are not always consistent in quality, which likely causes the FID scores to overfit the narrow reference corpus, making it unsuitable for evaluating graphic design quality.

TextDiffuser also yields higher CLIP scores.
We suspect this result is due to the text-spotting bias of the CLIP model, which tends to ``parrot'' the visual text embedding in images~\cite{lin2025parrot}.
That is, the CLIP score tends to prefer words over visual objects in the image, leading to a higher score for the word-only image without a meaningful background visual, which TextDiffusers tend to generate.

\begin{figure}[t]
\centering
\includegraphics[width=0.6\hsize]{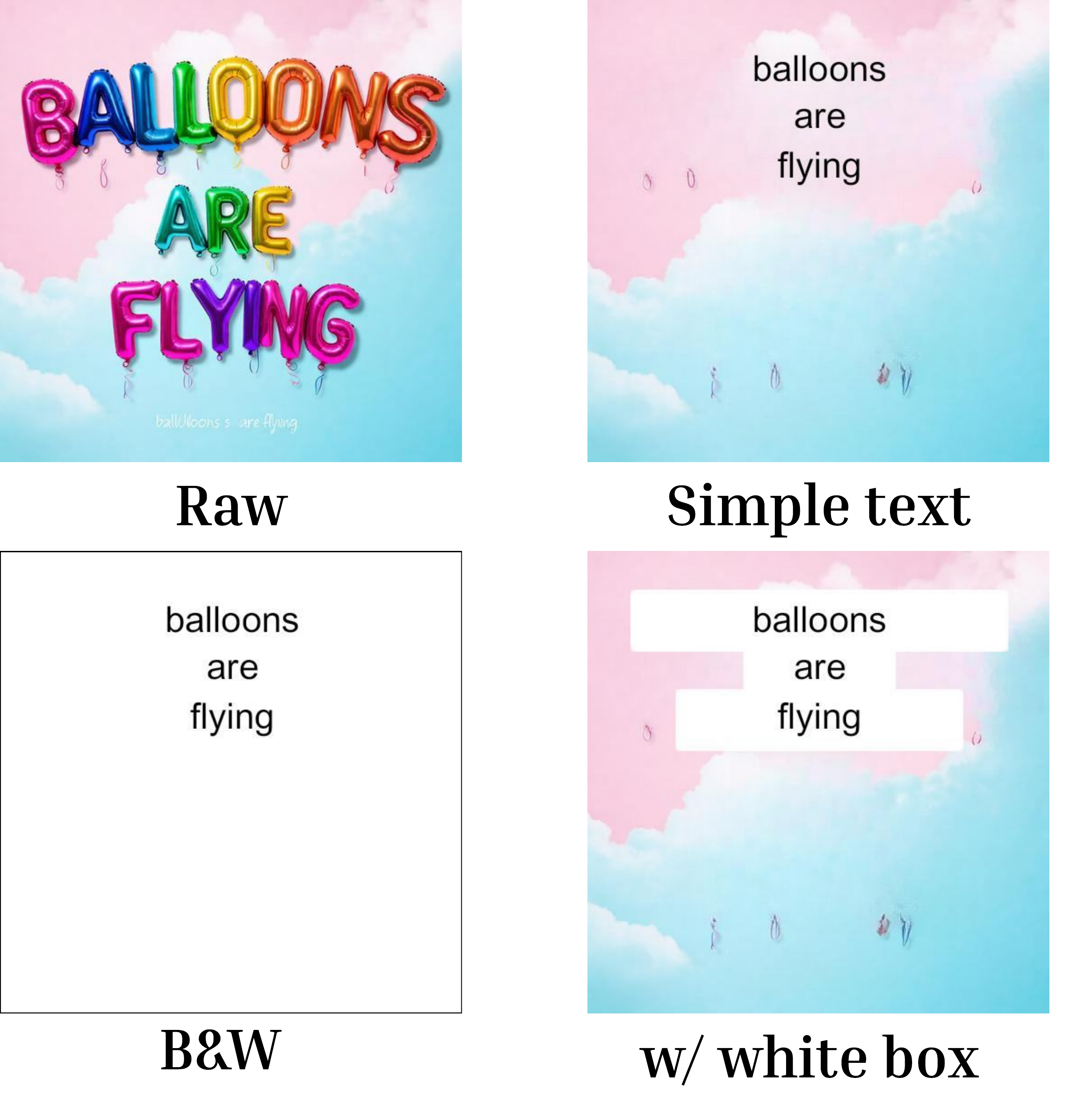}
\caption{Examples of naive baselines.}
\label{fig:naivebaselines}
\end{figure}

\begin{figure}[t] 
\centering
\includegraphics[width=\hsize]{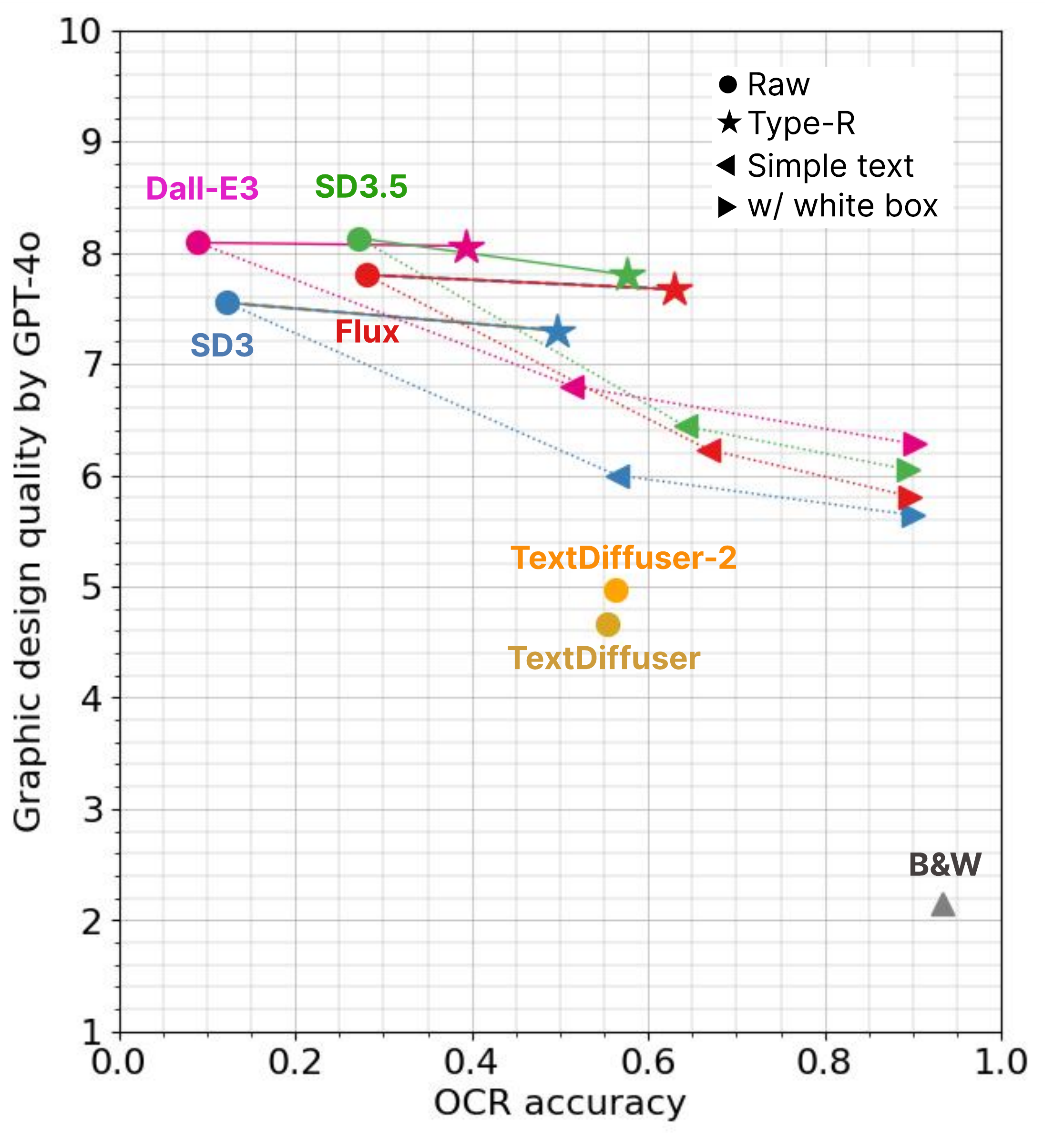}
\caption{Plot of the relation between OCR accuracy and graphic design quality by GPT. Raw represents the results of text-to-image generation with prompt augmentation.}
\label{fig:plot}
\end{figure}

\subsection{Additional Study}\label{sec:analysis}

\paragraph{Quality-accuracy Trade-off}
Here, we compare \ours{} with several baselines in terms of the trade-off between image quality and OCR accuracy.
In this study, we introduce the following backend text-to-image models.

\noindent\textbf{\ours{} w/ Dall-E 3} combines Dall-E 3~\cite{dalle3} with \ours{}.

\noindent\textbf{\ours{} w/ SD 3.5} combines Stable Diffusion 3.5~\cite{sd3dot5} with \ours{}.

We compare the raw output (\textbf{Raw}) from the backend text-to-image models and the output with \ours{}.
We also consider the following naive baselines (\cref{fig:naivebaselines}).
We heuristically arrange text boxes for all the naive baselines, ignoring the image content.

\noindent\textbf{Simple text} erases all texts in the generated image from the backend text-to-image models and uses it as a background.

\noindent\textbf{Simple text w/ white box} also removes all texts and white out the text box regions.

\noindent\textbf{B\&W} always renders text on a white background.

\cref{fig:plot} plots the trade-off between GPT score and OCR accuracy, which shows that naive baselines can achieve high OCR accuracy for much lower GPT score.
On the other hand, Flux and Dall-E 3 with \ours{} show much higher graphic design quality with a slightly degrading OCR accuracy, almost reaching the accuracy of the simple text rendering.
\ours{} exhibits robustness across multiple backend text-to-image models, as demonstrated in the plot where \ours{} effectively improves OCR accuracy for any backend models while maintaining high graphic design quality.

\paragraph{Qualitative Results}

\begin{figure}[t]
\centering
\includegraphics[width=\hsize]{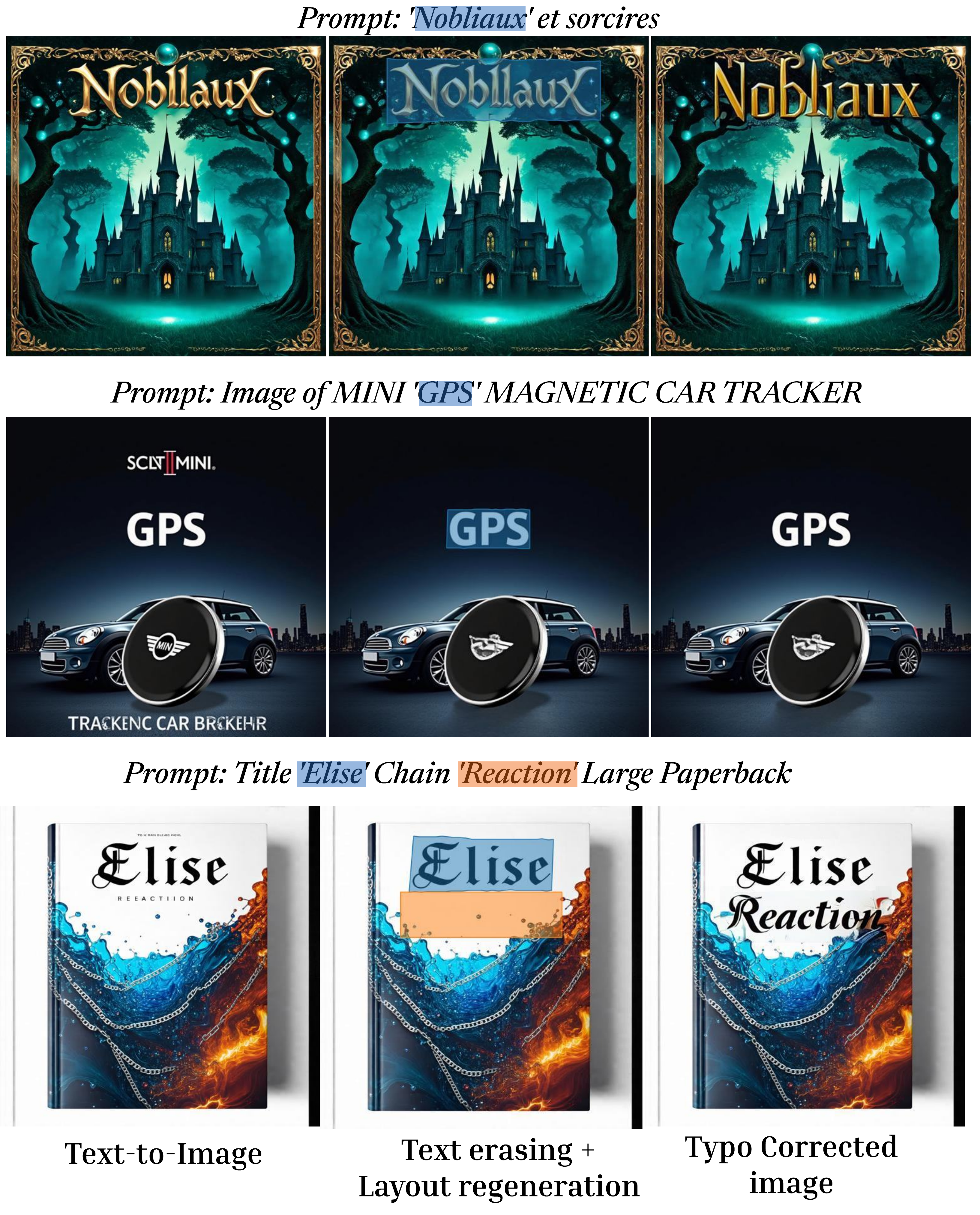}
\caption{Examples of generated images through \ours{}.}
\label{fig:examples}
\end{figure}

\Cref{fig:examples} shows a few generated images by \ours{} demonstrating how each external model improves text rendering accuracy.
In the first row, the text-editing model accurately modifies the text using a fancy font that aligns with the surrounding context. 
The second row presents an instance of \ours{} successfully removing unnecessary texts. 
The third row provides an example of removing small text and adding a missing text through layout planning, where the inserted new text implicitly utilizes surrounding font styles to achieve a seamless and natural appearance.

\begin{figure*}[t]
\centering
\includegraphics[width=\hsize]{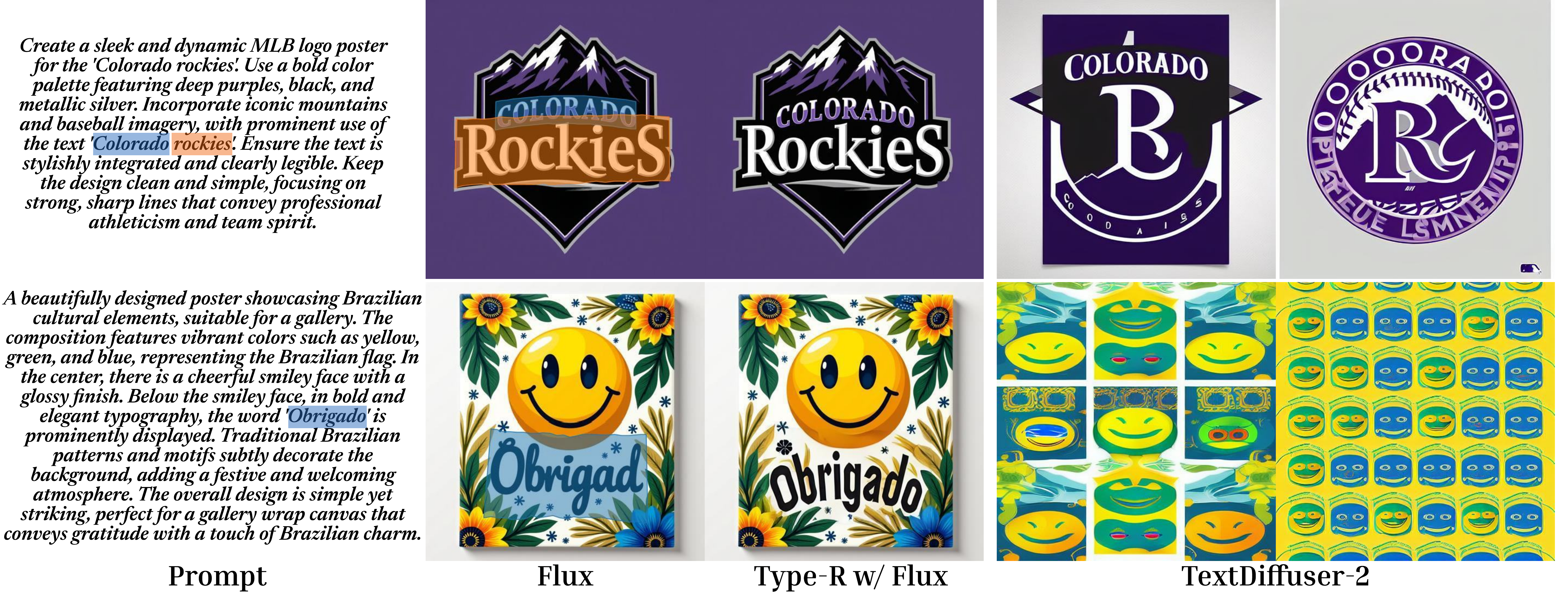}
\caption{Generated images from long prompts by Flux, Flux w/ \ours, and TextDiffuser-2.}
\label{fig:longprompt}
\end{figure*}

Since \ours{} is independent of the backend text-to-image models, we can take advantage of the state-of-the-art models.
\Cref{fig:longprompt} presents examples with very long prompts, showing that the latest text-to-image model, Flux, accurately captures detailed information from these prompts, whereas TextDiffuser-2 generates corrupted images when handling such prompts, as CLIP is unable to process tokens exceeding 77.

\paragraph{Longer Iterations}
\label{sec:iterationeffect}

\begin{figure}[t]
\centering
\includegraphics[width=\hsize]{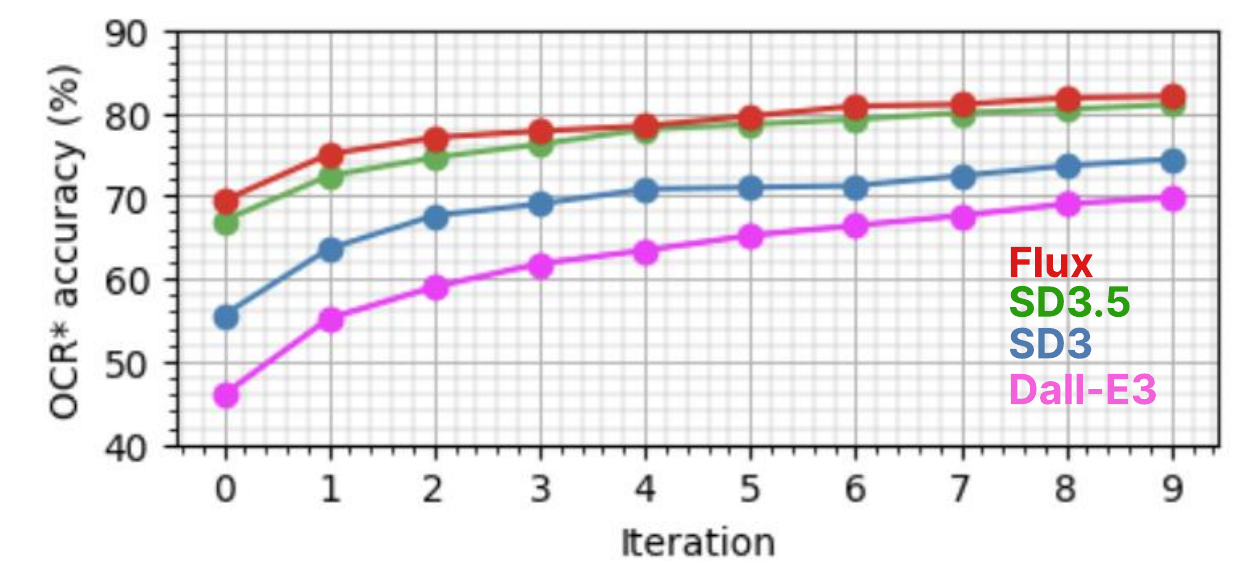}
\caption{Relationship in the typo correction iteration and OCR* accuracy. Note that OCR* accuracy is computed by OCR models in \ours{} which is distinct from the OCR model for evaluation.}
\label{fig:iteration}
\end{figure}

The final typo correction step iteratively refines misspelled words.
We show the improvement over iterations in \cref{fig:iteration} for several backend models.
Our typo correction approach consistently improves OCR accuracy as we repeat the iterations.
Our approach achieves 80 \% of the total improvement within four iterations independent of the backend model.

\subsection{Ablation Study}
\paragraph{\ours{} Pipeline Modules}
\cref{tab:modules} summarizes the ablation results for the modules in \ours{} pipeline: augmentation for prompt, layout correction (text erasing and layout regeneration), and typo correction.
We observe that each module effectively improves OCR accuracy with minimal negative impact on graphic scores by GPT.
Layout correction significantly improves Flux's OCR accuracy and further boosts OCR performance in combination with typo correction.
The comparison between Flux and SD3 suggests Flux exhibits lower rates of glyph corruption.

\begin{table}[t]
  \centering
\caption{Ablation study for \ours{} pipeline modules.} \label{tab:modules}
  \begin{tabular}{@{}cccccccc@{}}
    \hline
    \multirow{2}{*}{T2I}
    & \multirow{2}{*}{Aug} &
    \multicolumn{2}{c}{Correction}
    & \multirow{2}{*}{Graphic $\uparrow$}
    & \multirow{2}{*}{OCR $\uparrow$}
    \\
    & & Layout
    & Typo& \\ 
    \hline
  \multirow{5}{*}{SD3~\cite{esser2024scaling}}&   &  &  & 6.77 & 8.6 \\
   &  \checkmark &  &  & 7.55 & 12.2  \\
   &  \checkmark & \checkmark & & 7.42 & 27.4  \\
   &  \checkmark &  & \checkmark & 7.53 & 20.4  \\
   &  \checkmark & \checkmark & \checkmark & 7.27 & 50.0  \\ \hline
  \multirow{5}{*}{Flux~\cite{flux1dev}} &   &  &  & 6.79 & 14.4 \\
   &  \checkmark &  &  & 7.80 & 28.0   \\
   &  \checkmark & \checkmark & & 7.83 & 53.8  \\
   &  \checkmark &  & \checkmark & 7.73 & 29.4  \\
   &  \checkmark & \checkmark & \checkmark & 7.65 & 63.0  \\
    \hline
  \end{tabular}
\end{table}

\paragraph{OCR Models}

We summarize the effect of replacing the backend OCR models in \cref{tab:ocr}.
Both detection and recognition models impact OCR accuracy, with detection model performance changes having a more substantial effect than those of the recognition model.
We observe that CRAFT~\cite{craft} frequently merges words separated by spaces into a unit, leading to a notable performance drop, as \ours{} assumes that spaces are excluded from texts and does not account for this issue.
DeepSolo~\cite{deepsolo} achieves the highest performance.

For text recognition, improvements in OCR accuracy do not always align with scene text recognition performance.
While CLIP4STR-B*~\cite{clip4str} is one of the state-of-the-art scene text recognition models, the OCR accuracy of CLIP4STR-B* is lower than the OCR accuracy of Baek \etal~\cite{baek2019wrong} and Paddle~\cite{paddle}.
Paddle achieves the highest performance, likely because AnyText~\cite{anytext}, trained with a loss function that incorporates the Paddle model, tends to generate words that are more easily recognized by Paddle.
This suggests that compatibility between the text editing and recognition models is essential for effective correction.

\begin{table}[t]
  \centering
\caption{Comparing backend OCR models.} \label{tab:ocr}
  \begin{tabular}{@{}ccc@{}}
    \hline
    Detection 
    & Recognition
    & OCR $\uparrow$
    \\
    \hline
  Paddle~\cite{paddle} &  Paddle~\cite{paddle} & 57.4  \\
  CRAFT~\cite{craft} & Paddle~\cite{paddle}  & 47.2  \\
  Hi-SAM~\cite{hisam} & Paddle~\cite{paddle} & 60.8 \\
  Mask TextSpotter v3~\cite{mts} & Paddle~\cite{paddle}  & 61.0  \\
  DeepSolo~\cite{deepsolo} &  Paddle~\cite{paddle} & 63.0  \\
  DeepSolo~\cite{deepsolo} &  Baek et al.~\cite{baek2019wrong} & 62.0  \\
  DeepSolo~\cite{deepsolo} &  TrOCR~\cite{trocr} &  60.0 \\
  DeepSolo~\cite{deepsolo} &  CLIP4STR-B*~\cite{clip4str} & 60.8  \\
    \hline
  \end{tabular}
\end{table}

\paragraph{Text Erasing Methods}
We present the OCR accuracy when replacing the backend text erasing model in \cref{tab:eraser}.
In this study, we compare LaMa~\cite{suvorov2022resolution}, a generic inpainting model, and Garnet, a text-specific inpainting model.
Since inpainting models tend to be influenced by surrounding text regions, we optionally erase all text regions to prevent filling in letter-like artifacts and only focus on the background.
\Cref{tab:eraser} shows that LaMa performs well, and erasing all texts slightly improves the final OCR accuracy.
We observe that Garnet tends to leave outlines of texts when erasing text, and the outlines seem to be recognized as words in the OCR evaluation.

\begin{table}[t]
  \centering
\caption{Comparing backend text erasers.} \label{tab:eraser}
  \begin{tabular}{@{}ccc@{}}
    \hline
    Method & Erase all & OCR $\uparrow$ \\
    \hline
    \multirow{2}{*}{LaMa~\cite{suvorov2022resolution}} &  & 62.4 \\
     & \checkmark & 63.0 \\
     \hline
    \multirow{2}{*}{Garnet~\cite{lee2022surprisingly}} &  & 53.4 \\
     & \checkmark & 53.2 \\
    \hline
  \end{tabular}
\end{table}

\section{Limitation}
\ours{} depends on the performance of several external models and the base text-to-image models.
We observe that OCR models sometimes fail to correctly parse all texts generated by text-to-image models.
We also observe that AnyText~\cite{anytext} tends to generate unstable outlines of texts and degrading brightness of pixels from original images, which tends to cause incompatibility in composition.
Text-to-image models sometimes fail to render with detailed layout specifications and seem to suffer from uppercase and lowercase distinctions.
Although our method is applicable in any combination, we have to improve these backend models to resolve the failure cases.

There are other minor remaining problems.
\ours{} decomposes texts into words by the space.
There are texts whose reading order is meaningful, but \ours{} does not guarantee keeping the reading order of words in the generated image.
\ours{} has a relatively high computational requirement due to the stacking of multiple backend models in the pipeline, though it might be able to use the same backend in multiple stages to mitigate the issue; \eg, the test editing model can also erase texts.

\section{Conclusion}
In this paper, we propose an automatic typography correction method named \ours{}, which automatically detects and fixes typographic errors in the image generated by text-to-image models.
Through extensive experiments, we show that \ours{} effectively corrects typos while maintaining the quality of the generated images and outperforms baselines in terms of the quality-accuracy trade-off.
Our approach does not require training or fine-tuning and is readily pluggable to any existing text-to-image model.

\clearpage
\appendix
\renewcommand{\thesection}{\Alph{section}}
\setcounter{section}{0}
\makeatletter
\makeatother

\maketitlesupplementary

\section{GPT vs. Human Evaluation}
\label{sec:gptuserstudy}

\begin{figure}[t]
\centering
\includegraphics[width=1.0\hsize]{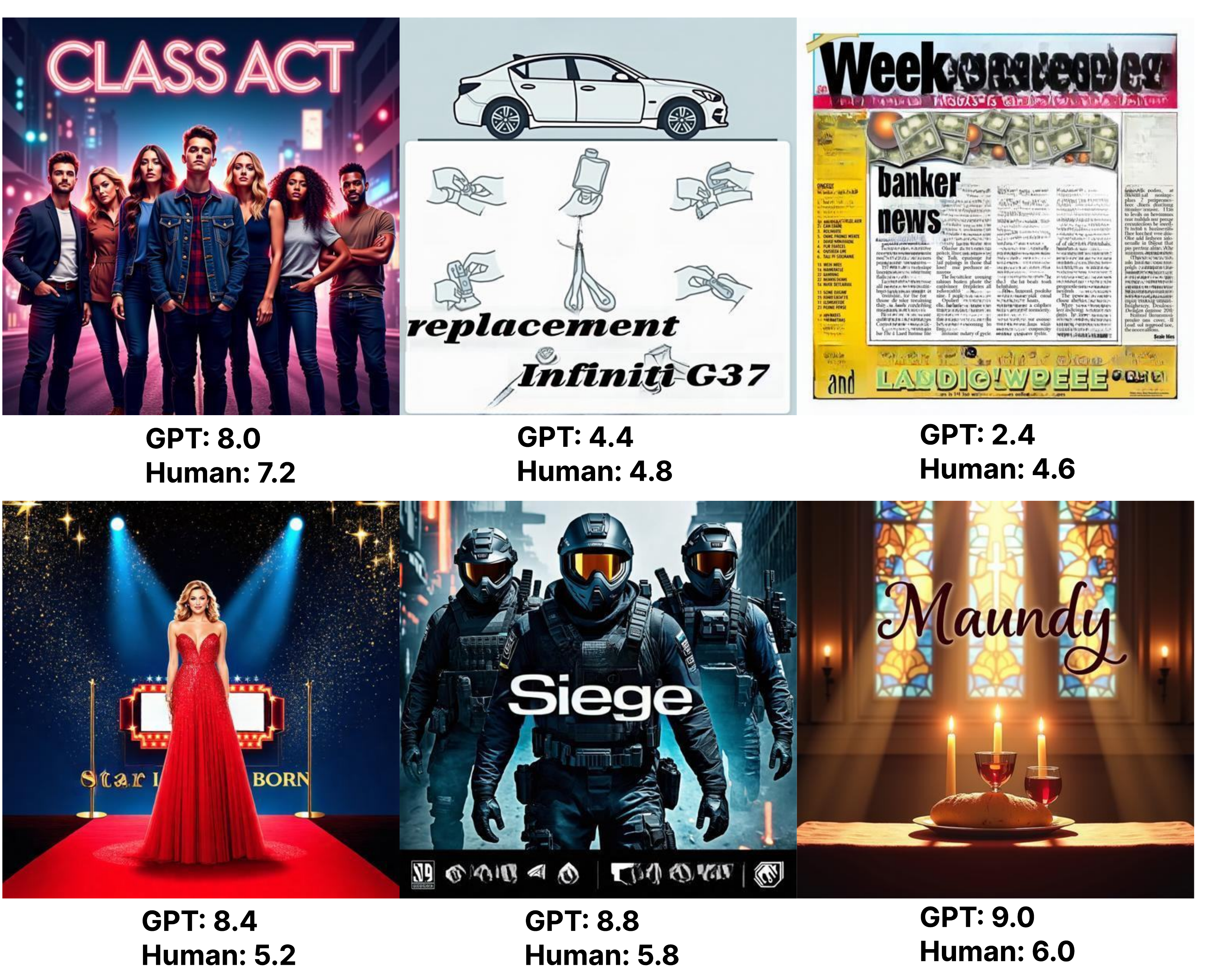}
\caption{GPT and human score examples.}
\label{fig:usscores}
\end{figure}

\Cref{fig:usscores} presents images generated by \ours{} along with their graphic design scores from both GPT and human evaluations.
We observe that GPT and human evaluations share common trends and unique tendencies.
The upper row highlights common trends.
GPT and human evaluators assign high scores to images with legible text and visually appealing elements, such as a design similar to a movie poster.
Conversely, both give low scores to images with corrupted layouts or distorted glyphs.
The bottom row illustrates unique trends.
GPT often assigns a high score to an image even when the image exhibits unstable outlines or unexplained artifacts, whereas humans typically rate such images lower.
Furthermore, humans tend to assign lower scores to an image with low-legibility text caused by poor contrast, while GPT is less sensitive to such issues.

\begin{figure*}[t]
\centering
\includegraphics[width=1.0\hsize]{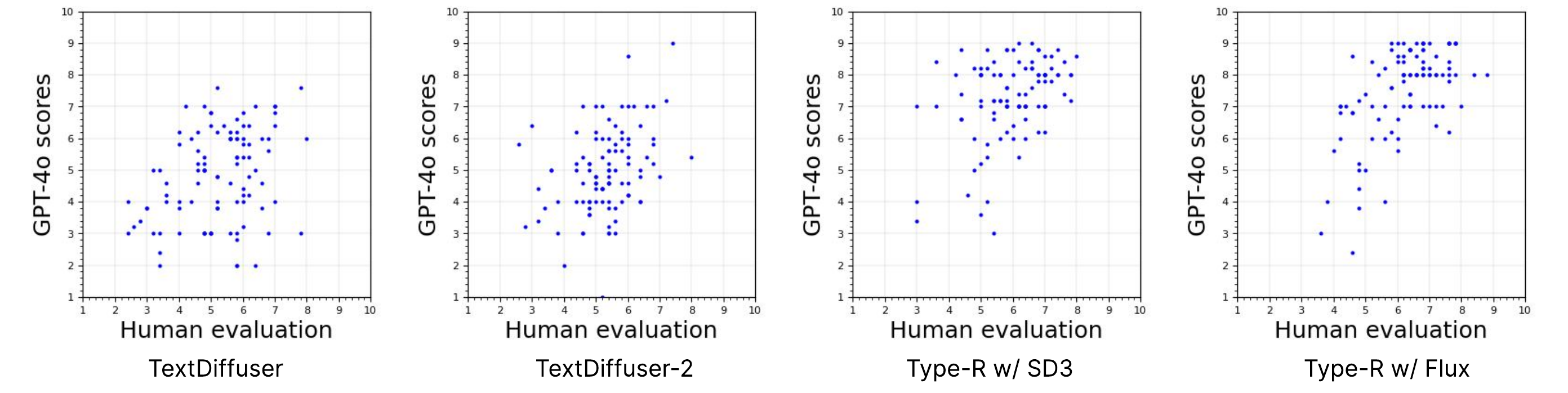}
\caption{Plots for graphic design scores by GPT and Human.}
\label{fig:usplots}
\end{figure*}

To better understand the differences, we present the GPT and human evaluation scores in \cref{fig:usplots}. 
Each plot includes 100 scores, averaged over five values from GPT and human evaluations.
The vertical axis represents GPT evaluation scores, while the horizontal axis corresponds to human evaluation scores.
The GPT and human evaluation scores show a rough correlation between images generated by different methods.
Notably, humans tend to avoid assigning extremely high or low scores.
On the plot of \ours{} w/ Flux, 56\% of the scores by GPT exceeds 8.0, compared to only 3\% by humans.
This suggests that humans may be better able to identify design defects or lack confidence in assigning high scores.
Despite some correlations between GPT and human evaluations, significant gaps remain.

\section{Voting Evaluation}
We replace the voting evaluation in TextDiffuser-2~\cite{textdiffuser2} with rating evaluation~\cite{jia2023cole, inoue2024opencole} due to its greater sensitivity to factors such as candidate combinations or the order in which candidates are presented.
Here, we use rating to eliminate the sensitivity associated with voting evaluation using a binary voting format.
\Cref{tab:voting} presents the results of the voting evaluation for TextDiffuser-2 and \ours{} w/ Flux, focusing on design quality, text-image matching, and text quality.
The evaluation scheme for text-image matching and text quality follows the approach used in TextDiffuser-2~\cite{textdiffuser2}.
For design quality evaluation, we replace ``text quality'' with ``design quality'' in the prompt used for text quality evaluation.
We observe that \ours{} w/ Flux is consistently preferred by GPT across all three perspectives.

\begin{table}[t]
  \centering
\caption{Binary voting evaluation for TextDiffuser-2 and \ours{} w/ Flux.} \label{tab:voting}
  \begin{tabular}{@{}lcc@{}}
    \hline
    Method & TextDiffuser-2~\cite{textdiffuser2} &  Type-R w/ Flux \\ 
    \hline
    Design Quality $\uparrow$ & 5.0 & \textbf{95.0} \\
    Matching $\uparrow$ & 29.1 & \textbf{70.9} \\
    Text Quality $\uparrow$  & 28.0  &  \textbf{72.0} \\
    \hline
  \end{tabular}
\end{table}

\begin{table*}[t]
  \centering
\caption{Grid search for the hyperparameter of filtering on the validation data. The rate means the proportion of text height to image height} \label{tab:filter}
  \begin{tabular}{@{}ccccccccccc@{}}
    \hline
    Rate \(\%\) 
    &1.0
    & 2.0
    & 3.0
    & 4.0
    & 5.0
    & 6.0
    & 7.0
    & 8.0
    & 9.0
    & 10.0
    \\
    \hline
    OCR $\uparrow$
    & 60.6 & 60.6 & 62.4 & \textbf{64.4} & 62.2 & 59.0 &  59.8 & 57.2 & 56.0 & 56.0 \\
    \hline
  \end{tabular}
\end{table*}

\begin{figure*}[t]
\centering
\includegraphics[width=\hsize]{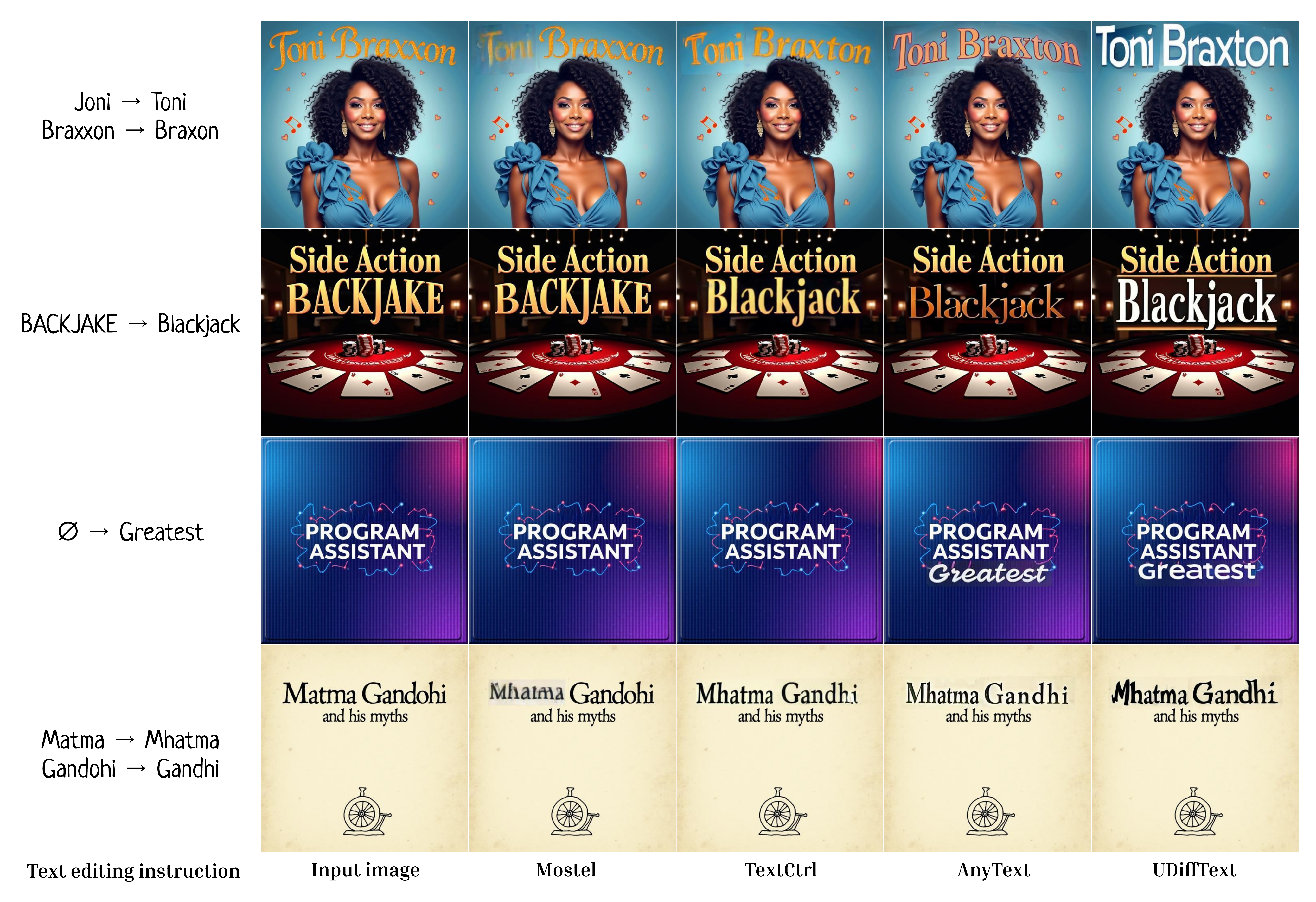}
\caption{Examples of typo correction across text editing methods.}
\label{fig:text_editing}
\end{figure*}

\section{Grid Search for Filtering}

We remove small texts based on the hyperparameter $\theta$, which represents the proportion of text height to image height.
We find this hyperparameter through a grid search.
\Cref{tab:filter} shows the grid search results for the filtering ratio.
$\theta = 4\%$ yields the best OCR accuracy, and we select this ratio accordingly.
Note that we assume the aspect ratio of the image is 1:1.

\begin{table}[t]
  \centering
\caption{Comparison with the approach proposed by Li \etal{}. * indicates some modules are replaced with the same modules in \ours{}.} \label{tab:wordmatching}
  \begin{tabular}{@{}ccc@{}}
    \hline
    Method & Graphic design score $\uparrow$ & OCR $\uparrow$ \\
    \hline
    Li \etal~\cite{Li2024first}* w/ Flux & 7.00  & 55.4\\
    Type-R  w/ Flux & \textbf{7.65} & \textbf{63.0} \\
    \hline
  \end{tabular}
\end{table}

\begin{table}[t]
  \centering
\caption{Ablation study for the text editor. * represents the methods partially used in not intended usage.} \label{tab:texteditor}
  \begin{tabular}{@{}ccc@{}}
    \hline
    Method & Graphic design score $\uparrow$ & OCR $\uparrow$ \\
    \hline
    Mostel*~\cite{mostel} & \textbf{7.88}  & 53.6 \\
    TextCtrl*~\cite{zeng2024textctrl} & 7.80 & 62.0 \\
    AnyText~\cite{anytext} & 7.65 & 63.0 \\
    UDiffText~\cite{zhao2025udifftext} & 7.66 & \textbf{67.8} \\
    \hline
  \end{tabular}
\end{table}

\section{Error Detection and Matching}

\ours{} leverages rendered texts from text-to-image models, whereas Li \etal~\cite{Li2024first} propose to erase all texts in a generated image to create a canvas for text rendering.
To validate the effects of both approaches, we prepare a method that removes all texts from generated images by text-to-image models, then generates text layouts through layout regeneration and creates text images using AnyText~\cite{anytext}, i.e., this method generates text without error detection and word matching.
\Cref{tab:wordmatching} shows the results of the ablation study for error detection and word matching.
Skipping error detection and word matching significantly reduces OCR accuracy and causes a notable decrease in graphic design scores by GPT.
These results highlight the importance of fully utilizing the initially rendered texts from state-of-the-art text-to-image models.

\section{Text Editing Methods}

\ours{} relies on text editing methods, and we investigate how the choice of text editing method affects the results.
In particular, we examine four text editing methods: Mostel*~\cite{mostel}, TextCtrl*~\cite{zeng2024textctrl}, AnyText~\cite{anytext}, and UDiffText~\cite{zhao2025udifftext}.
Mostel* and TextCtrl* are representative text editing methods that replace text in an image with a specified alternative while preserving the text's style.
We adapt these methods iteratively to achieve multiple text edits in an image through \ours{}. When drawing text to blank space using these text editing methods, we provide a background image of blank space even though this is not the intended use.
AnyText and UDiffText are inpainting-based models that can both edit existing text and generate text for blank spaces in an inpainting manner.
AnyText can generate multiple texts in a single inference step, whereas UDiffText generates one text per inference step.
When using UDiffText in \ours{}, we adapt it iteratively for multiple text generations, similar to how we handle text editing methods.

\Cref{tab:texteditor} shows the GPT scores and OCR accuracy, while \cref{fig:text_editing} displays the qualitative results.
We observe that the OCR accuracy of Mostel* is the lowest.
The first row in \cref{fig:text_editing} shows that text editing with Mostel* results in blurred.
Mostel* frequently generates a text that remains unreadable by the scene text recognition model, with the target text often left unchanged, as shown in the second row of the figure \cref{fig:text_editing}.
This issue likely arises from the limited robustness of Mostel*, which is based on GAN modeling.
TextCtrl provides clearer results compared to Mostel*, and its OCR accuracy is close to that of AnyText.
However, TextCtrl consistently fails to render text for blank spaces, as shown in the third row of \cref{fig:text_editing}, due to its text-editing approach.
AnyText and UdiffText successfully generate text for both text-rendered regions and blank spaces.
UdiffText outperforms AnyText in terms of OCR accuracy, and the graphic design scores by GPT for both methods are close.
However, UdiffText tends to exhibit unstable outlines, as seen in the fourth row of \cref{fig:text_editing}.
As discussed in \cref{sec:gptuserstudy}, GPT evaluations overlook small artifacts, so the instability of the outlines does not seem to impact the graphic design score.
We believe that achieving a balance between high text rendering accuracy and stable glyph outlines is crucial for effectively retouching typos.

\begin{figure*}[t]
\centering
\includegraphics[width=\hsize]{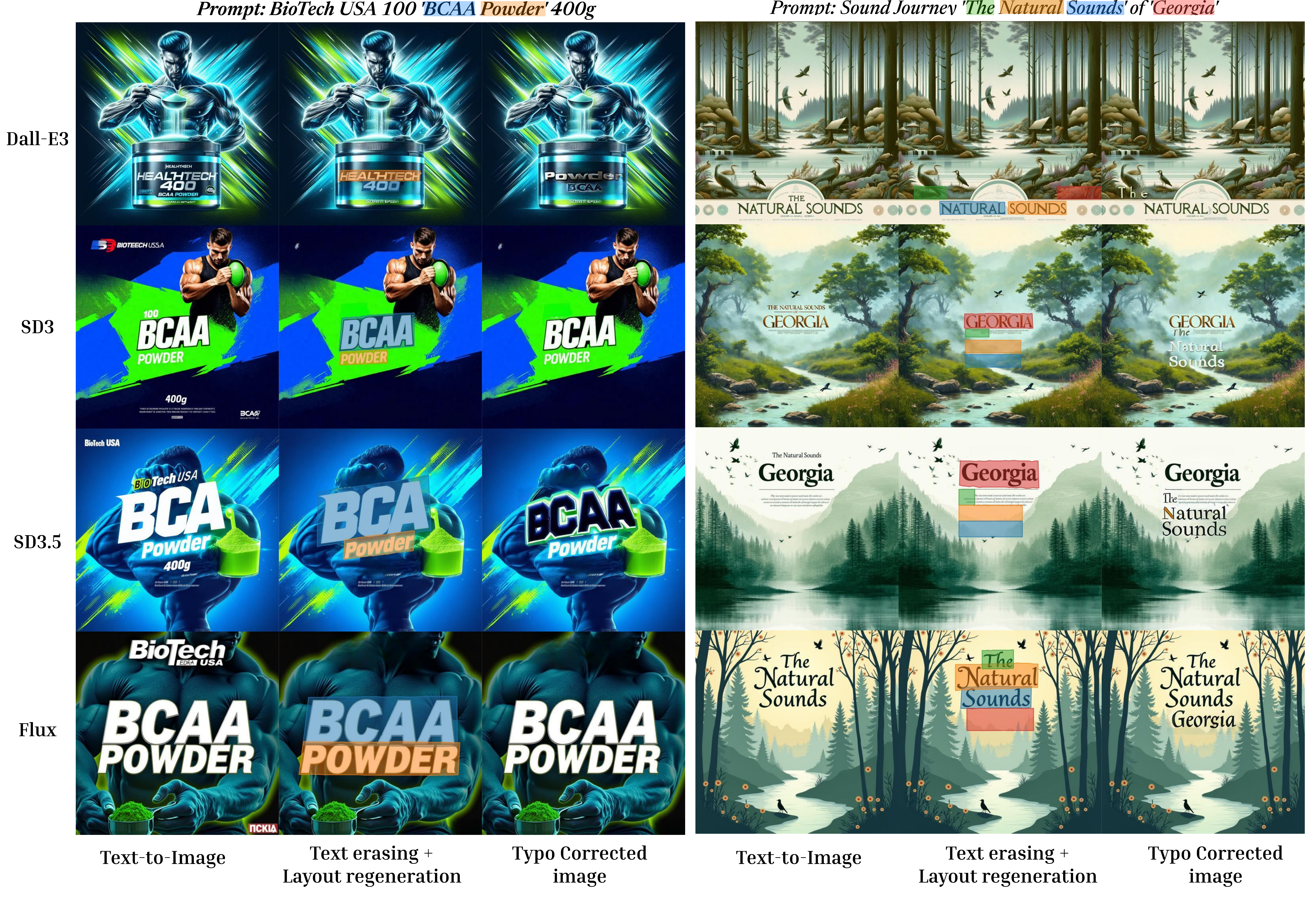}
\caption{Typo retouched images by \ours{} across text-to-image models.}
\label{fig:t2i}
\end{figure*}

\section{Number of Error Correction through \ours{}}

To understand why raw text-to-image models perform poorly on OCR accuracy and how \ours{} addresses these issues, we present the change in the number of errors in \cref{tab:wordstats}.
Note that the granularity of the errors is at the word level, and we compute the statistics on a subset of the MARIO-Eval benchmark, consisting of 500 prompts.
The statistics represent pseudo-errors, which are computed through the OCR models used in \ours{}.

We observe that the number of detected words is more than twice the number of words specified to render in the prompts across text-to-image models.
While text-to-image models rarely fail to render the required number of words in generated images, they often produce additional unintended words.
This suggests that the training data for text-image pairs is not entirely accurate.
\ours{} resolves the errors, although layout regeneration sometimes fails due to corruption in the Vision-Language Model (VLM) output.
Although typo correction does not completely eliminate errors, it reduces the error rate by half.

\section{Layout Regeneration}
\begin{table*}[t]
  \centering
\caption{Statistics for the number of words through \ours{}. The total number of prompt words is 1426.} \label{tab:wordstats}
  \begin{tabular}{@{}cccccc@{}}
    \hline
    Text-to-image
    & Detected words
    & Surplus words
    & Lack words
    & Typo words
    & Typo corrected words
    \\
    \hline
    DallE3 & 3449 & 2243 & 572 & 483 & 252 \\

    SD3 
     & 3307 & 1976 & 334 &  397 & 198 \\
    SD3.5 
     & 2963 & 1673 & 341 &  244 & 134 \\
    Flux 
     & 3589 & 2255 & 170 &  252 & 132 \\
    \hline
  \end{tabular}
\end{table*}

\begin{table}[t]
  \centering
\caption{Ablation study for layout regeneration.} \label{tab:lp}
  \begin{tabular}{@{}cc@{}}
    \hline
    Method & OCR $\uparrow$ \\
    \hline
    w/o layout regeneration & 58.6 \\
    w/ layout regeneration & \textbf{63.0} \\
    \hline
  \end{tabular}
\end{table}

\Cref{tab:lp} shows the effect of layout regeneration on OCR accuracy.
The performance improvement in OCR accuracy due to layout regeneration is approximately five percentage points.
Although the number of errors eliminated by layout regeneration is fewer than text erasing, the effect of layout regeneration is still significant.

\begin{figure}[t]
\centering
\includegraphics[width=\hsize]{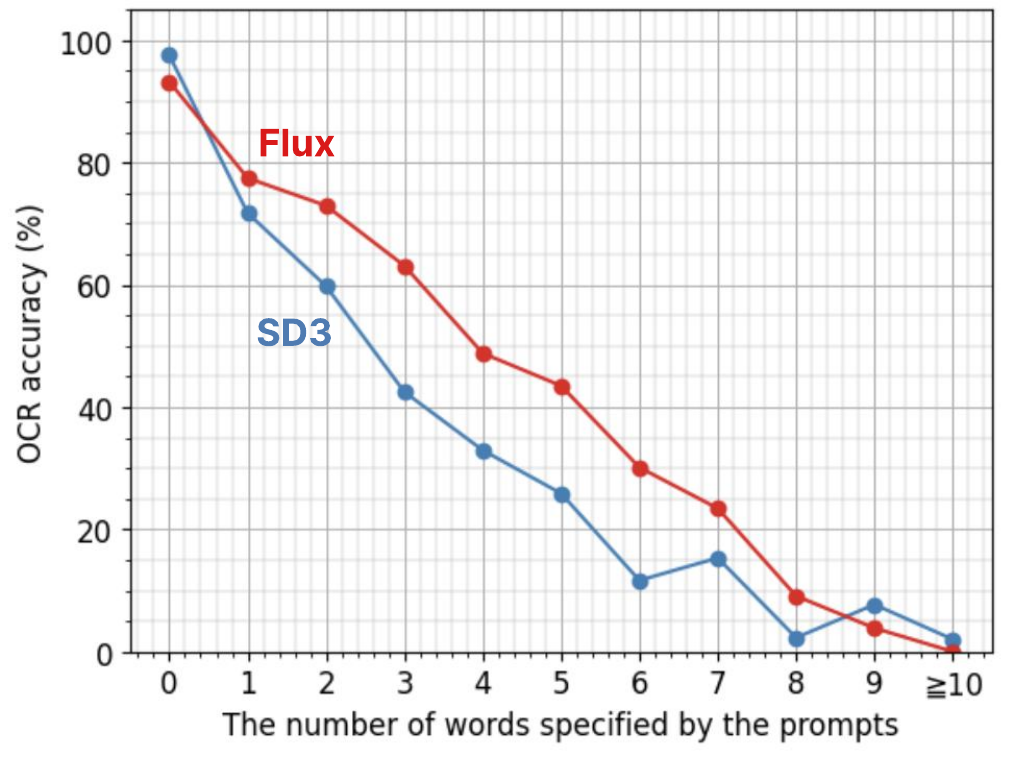}
\caption{The plots of capacity for handling texts through \ours{}.}
\label{fig:wordnum_ocracc}
\end{figure}

\begin{figure}[t]
\centering
\includegraphics[width=\hsize]{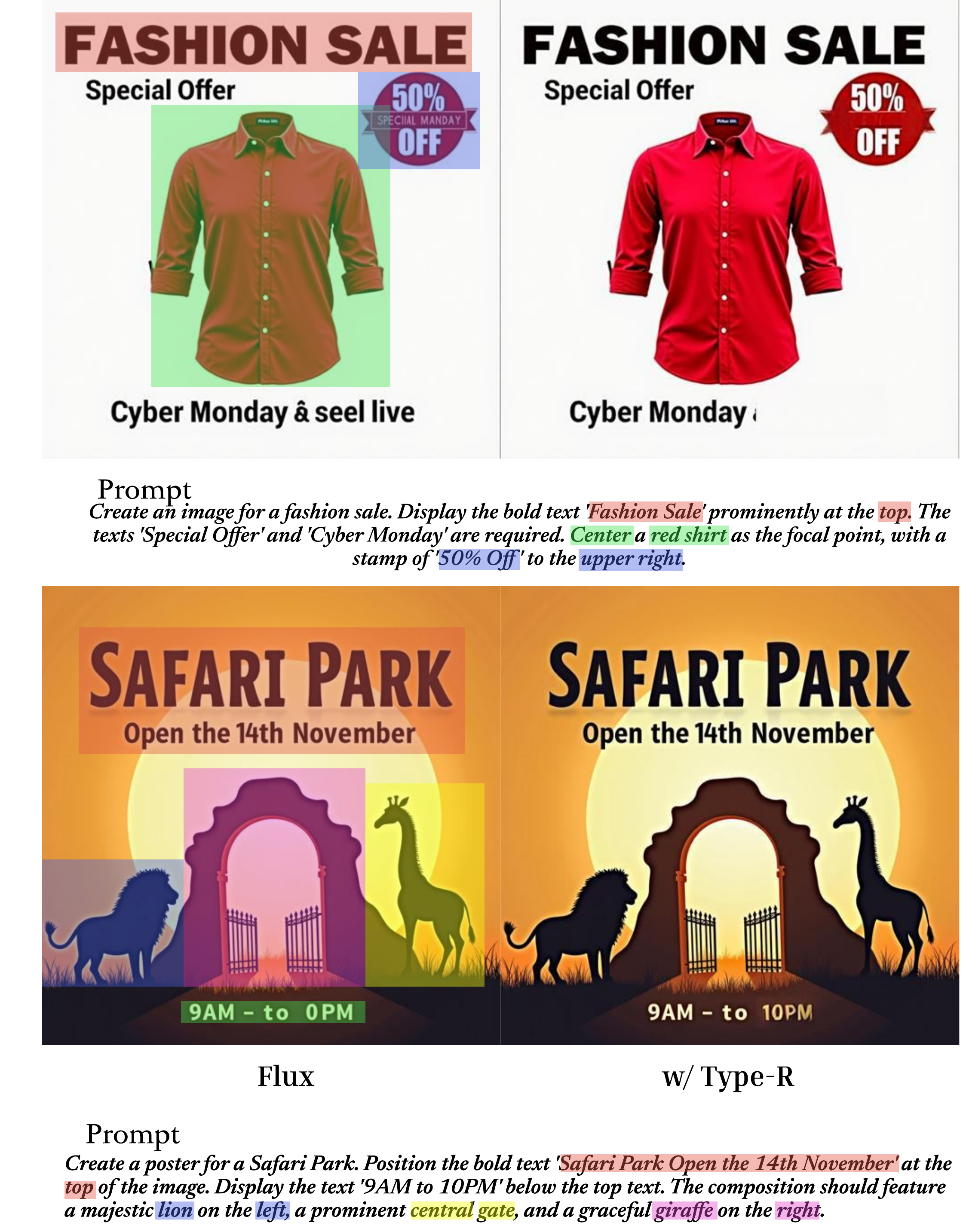}
\caption{Examples of results from \ours{} using rough layout specifications with prompts.}
\label{fig:roughlayout}
\end{figure}

\begin{figure*}[t]
\centering
\includegraphics[width=\hsize]{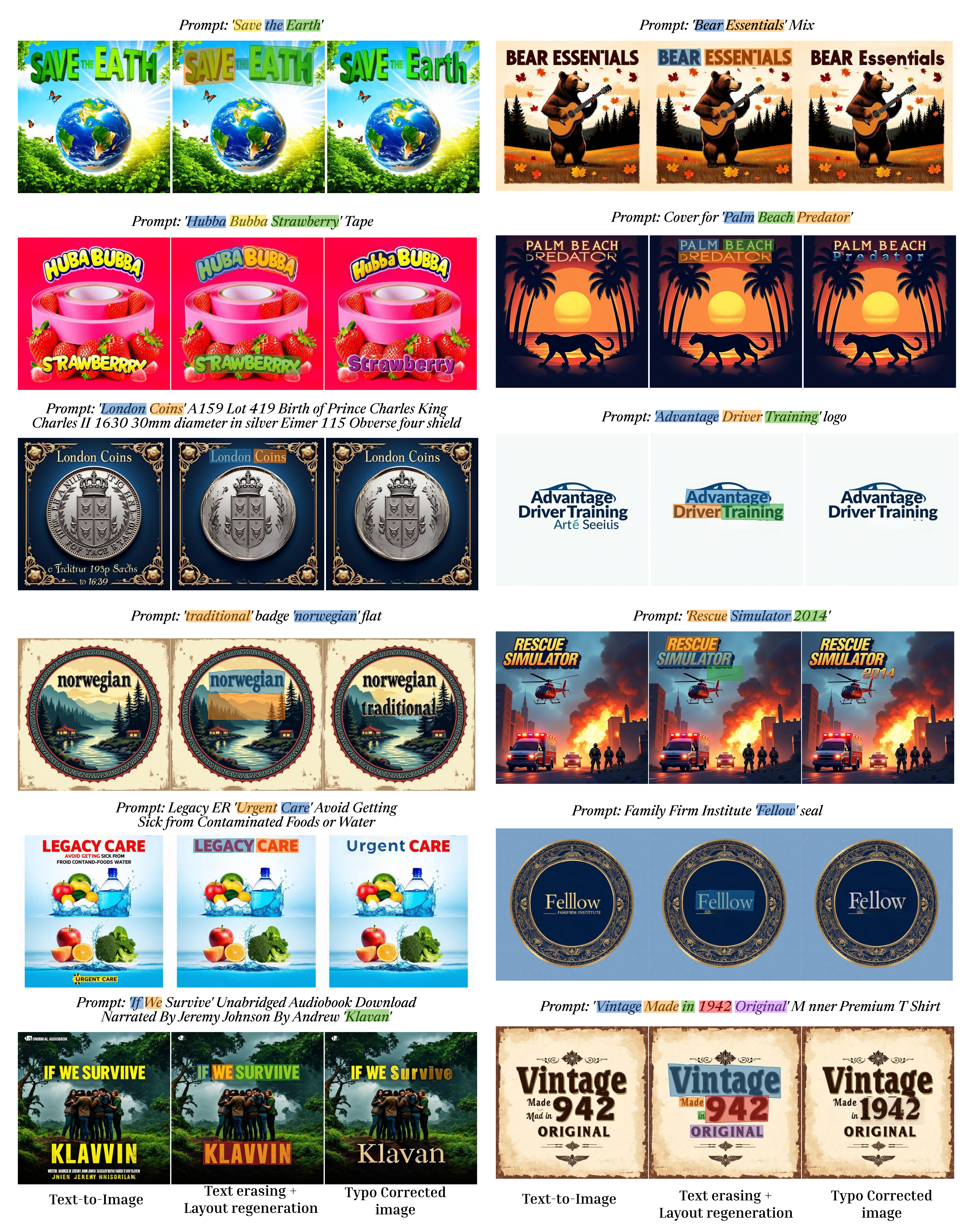}
\caption{Examples of generated images through \ours{}.}
\label{fig:moreexamples}
\end{figure*}

\section{Text-to-Image Models}
The quality of generated images by text-to-image models plays a crucial role in \ours{}.
\Cref{tab:t2i} shows the performance of OCR accuracy and graphic scores by GPT across various text-to-image models through \ours{}.
While the differences in graphic scores by GPT are relatively small, there are significant performance gaps in OCR accuracy across the text-to-image models.
Flux~\cite{flux1dev} achieves the best OCR accuracy, while Dall-E3~\cite{dalle3} achieves the worst score.
The number of errors observed in layout regeneration likely correlates to the ease of improving OCR accuracy through \ours{}.
We expect that error correction through text erasing tends to be effective, while layout regeneration often leads to corruption, meaning that error correction via layout regeneration is less accurate than text erasing.

\Cref{fig:t2i} provides examples of retouching results across text-to-image models using \ours{}. The left samples showcase typical cases of error correction through text erasing, while the right samples demonstrate cases with more layout regeneration, particularly with images generated by Dall-E 3, SD3~\cite{esser2024scaling}, and SD3.5~\cite{sd3dot5}.
These images, which require rendering many words through layout regeneration, often result in poor visual quality.
We expect that Flux involves less layout regeneration due to fewer issues with missing words in text rendering, contributing to its higher OCR accuracy than other models.

\section{Capacity for Handling Texts}

We investigate the capacity for handling text using \ours{}.
\Cref{fig:wordnum_ocracc} shows the plots depicting the relationship between OCR accuracy and the number of words to render in the generated images for \ours{} with Flux and \ours{} with SD3.
We observe that OCR accuracy decreases almost linearly as the number of words specified in the prompts increases.
Note that this plot is generated using the full set of the Mario-Eval benchmark.

\begin{table}[t]
  \centering
\caption{Ablation study for the text-to-image models.} \label{tab:t2i}
  \begin{tabular}{@{}ccc@{}}
    \hline
    Method & Graphic design score $\uparrow$ & OCR $\uparrow$ \\
    \hline
    Dall-E3 & \textbf{8.06}  & 39.4 \\
    SD3 & 7.27 & 49.6 \\
    SD3.5 & 7.80 & 57.6 \\
    Flux & 7.65 & \textbf{63.0} \\
    \hline
  \end{tabular}
\end{table}

\section{Additional Qualitative Results}

We present additional qualitative examples in this section.
\Cref{fig:roughlayout} shows examples that demonstrate the benefits of state-of-the-art text-to-image models.
The latest model, Flux, accurately captures the rough layout information from the prompts.
We highlight the words in the same color to show the pairs of rough layout information and their corresponding target objects in the prompt.
We confirm that Flux can reflect the rough layout information in the generated images, effectively positioning objects and text as specified in the prompt.

We further demonstrate how \ours{} improves text rendering accuracy through each external model in \cref{fig:moreexamples}.
The first and second rows show \ours{} successfully correcting typos by editing texts through matching similar words.
The third row presents an instance of Type-R effectively removing unnecessary text.
The fourth row provides examples of missing text through layout regeneration, where the inserted text seamlessly integrates with the surrounding font styles to create a natural and coherent appearance.
The fifth and sixth rows show successful cases combining text erasing and typo correction.
\section{Prompts}
We use GPT-4o~\cite{gpt4o} for prompt augmentation, layout regeneration, and evaluation. In this section, we provide the prompts used for each of these tasks.

\subsection{Prompt augmentation}

In prompt augmentation, we augment prompts using GPT-4o. We assume that the texts to be rendered are enclosed in either quotation marks or double quotation marks and validate that the augmented prompts contain the enclosed texts correctly. If the validation fails, we retry the augmentation process up to five times.
If the retry limit is reached (five attempts), we replace the prompt with a template:
"Draw a picture about [\texttt{prompt}] with the large text "[\texttt{target texts}]",
where \texttt{prompt} refers to the original prompt with quotations and double quotations removed, and \texttt{target texts} is the joined text to be rendered, separated by spaces.
When augmenting prompts, we submit the following system prompt:

\begin{intentionpromptaug}
You are an excellent autonomous AI Assistant.
Given a short prompt, generate a concise yet expressive augmented prompt, which will be used as the input of text-to-image models.
The augmented prompt should at least follow some rules:

(1) It should include references to data domains such as advertisements, notes, posters, covers, memes, logos, and books.

(2) It should mention as many features as possible, such as objects and their composition, colors, and overall atmosphere.

(3) All text enclosed in single or double quotes (e.g., 'Michael Jordan') should be displayed legibly in the image, while any other text should not be included. The augmented prompt must specify all text intended for display by using either single or double quotes.

(4) Quotation marks are solely for indicating text to be drawn in the image and should not be used for any other purposes, such as possessives.

(5) A simpler design organization is preferable.

(6) For any other text, interpret the context from the short input and feel free to expand where appropriate.
\end{intentionpromptaug}

\subsection{Layout regeneration}

To obtain layout information, we ask GPT to plan layouts for each missing word in the image, along with the layout information for valid text boxes in JSON format.
We submit the following system prompt with an input image and missing keywords:

\begin{intentionpromptregeneration}
You are an excellent autonomous AI Assistant. Please plan the layout for a list of keywords, given the image and layout information on already printed texts. Note that the canvas size is 128x128.
The output should be formatted as a JSON instance that conforms to the JSON schema below.

As an example, for the schema {"properties": {"foo": {"title": "Foo", "description": "a list of strings", "type": "array", "items": {"type": "string"}}}, "required": ["foo"]}
the object {"foo": ["bar", "baz"]} is a well-formatted instance of the schema. The object {"properties": {"foo": ["bar", "baz"]}} is not well-formatted.

Here is the output schema:

```

{"\$defs": \{"Element": \{"properties": \{"word": \{"title": "Word", "type": "string"\}, "width": \{"maximum": 128, "minimum": 1, "title": "Width", "type": "integer"\}, "height": \{"maximum": 128, "minimum": 1, "title": "Height", "type": "integer"\}, "left": \{"maximum": 127, "minimum": 0, "title": "Left", "type": "integer"\}, "top": \{"maximum": 127, "minimum": 0, "title": "Top", "type": "integer"\}\}, "required": ["word", "width", "height", "left", "top"], "title": "Element", "type": "object"\}\}, "properties": \{"elements": \{"default": [], "items": \{"\$ref": "\#/\$defs/Element"\}, "title": "Elements", "type": "array"\}\}\}

```

Below are some typical examples

Current layout: \{"elements": [\{"word": "Hello", "width": 64, "height": 16, "left": 32, "top": 32}]\} Input keywords: ["world!"] Output: \{"elements": [\{"word": "world!", "width": 64, "height": 16, "left": 32, "top": 48\}]\}

\end{intentionpromptregeneration}

\subsection{Evaluation}
To rate generated images based on graphic design quality, we use the following system prompt:

\begin{intentionprompt}

You are an autonomous AI Assistant who aids designers by providing insightful, objective, and constructive critiques of graphic design projects. 

Your goals are: Deliver comprehensive and unbiased evaluations of graphic designs based on established design principles and industry standards. Maintain a consistent and high standard of critique. Please abide by the following rules: 

Strive to score as objectively as possible. Grade seriously. A flawless design can earn 10 points, a mediocre design can only earn 7 points, a design with obvious shortcomings can only earn 4 points, and a very poor design can only earn 1-2 points. The overall looks of the image please consider factors such as the design, layout, typography, rendering quality, color scheme, harmony, innovation, and originality in your evaluation. Keep your reasoning concise when rating, and describe it as briefly as possible. \"explanation\" should start with \"Let's think step by step\"."
\end{intentionprompt}

To rate generated images based on text-image matching, we use the following system prompt
\begin{intentionprompt}
You are an autonomous AI Assistant who aids designers by providing insightful, objective, and constructive critiques of graphic design projects. 

Your goals are: Deliver comprehensive and unbiased evaluations of graphic designs based on established design principles and industry standards. Maintain a consistent and high standard of critique. Please abide by the following rules: 

Strive to score as objectively as possible. Grade seriously. The content should be not only relevant to its purpose but also engaging for the intended audience, effectively communicating the intended message.  A score of 10 means the content resonates with the target audience, aligns with the design's purpose, and enhances the overall message. A score of 1 indicates the content is irrelevant or does not connect with the audience. Keep your reasoning concise when rating, and describe it as briefly as possible. \"explanation\" should start with \"Let's think step by step\"."
\end{intentionprompt}

{
    \small
    \bibliographystyle{ieeenat_fullname}
    \bibliography{main}
}

\end{document}